\documentclass[10pt,twocolumn,letterpaper]{article}

\usepackage{cvpr}              

\usepackage[dvipsnames]{xcolor}

\definecolor{cvprblue}{rgb}{0.21,0.49,0.74}
\usepackage[pagebackref,breaklinks,colorlinks,citecolor=cvprblue]{hyperref}
\usepackage{booktabs}
\usepackage{multirow}
\usepackage{graphicx}
\usepackage{pifont}
\usepackage{diagbox}
\usepackage[accsupp]{axessibility}

\newcommand{\ieno}{\textit{i}.\textit{e}.}
\newcommand{\egno}{\textit{e}.\textit{g}.} 

\def\paperID{4659} 
\def\confName{CVPR}
\def\confYear{2024}

\title{Text Grouping Adapter: Adapting Pre-trained Text Detector for Layout Analysis}

\author{Tianci Bi$^1$\footnotemark[1] \quad Xiaoyi Zhang$^2$ \quad Zhizheng Zhang$^2$ \quad Wenxuan Xie$^2$ \quad 
Cuiling Lan$^2$ \\ \quad Yan Lu$^2$\footnotemark[3] \quad Nanning Zheng$^1$\footnotemark[3] \\
$^1$ IAIR, Xi’an Jiaotong University \\
$^2$ Microsoft Research Asia \\
\tt \small tiancibi@stu.xjtu.edu.cn, \\
\tt \small {xiaoyizhang, zhizzhang, wenxie, culan, yanlu}@microsoft.com, nnzheng@mail.xjtu.edu.cn}

\begin{document}

\maketitle
\footnotetext[1]{Work done during the internship at Microsoft Research Asia.}
\footnotetext[3]{Corresponding authors.}

\begin{abstract}

Significant progress has been made in scene text detection models since the rise of deep learning, but scene text layout analysis, which aims to group detected text instances as paragraphs, has not kept pace. Previous works either treated text detection and grouping using separate models, or train a model from scratch while using a unified one. All of them have not yet made full use of the already well-trained text detectors and easily obtainable detection datasets. In this paper, we present Text Grouping Adapter (TGA), a module that can enable the utilization of various pre-trained text detectors to learn layout analysis, allowing us to adopt a well-trained text detector right off the shelf or just fine-tune it efficiently. Designed to be compatible with various text detector architectures, TGA takes detected text regions and image features as universal inputs to assemble text instance features. To capture broader contextual information for layout analysis, we propose to predict text group masks from text instance features by one-to-many assignment. Our comprehensive experiments demonstrate that, even with frozen pre-trained models, incorporating our TGA into various pre-trained text detectors and text spotters can achieve superior layout analysis performance, simultaneously inheriting generalized text detection ability from pre-training. In the case of full parameter fine-tuning, we can further improve layout analysis performance.
\end{abstract}

\vspace{-3mm}
\section{Introduction}
\vspace{-1mm}

\begin{figure}
    \centering
    \includegraphics[width=1.0\linewidth]{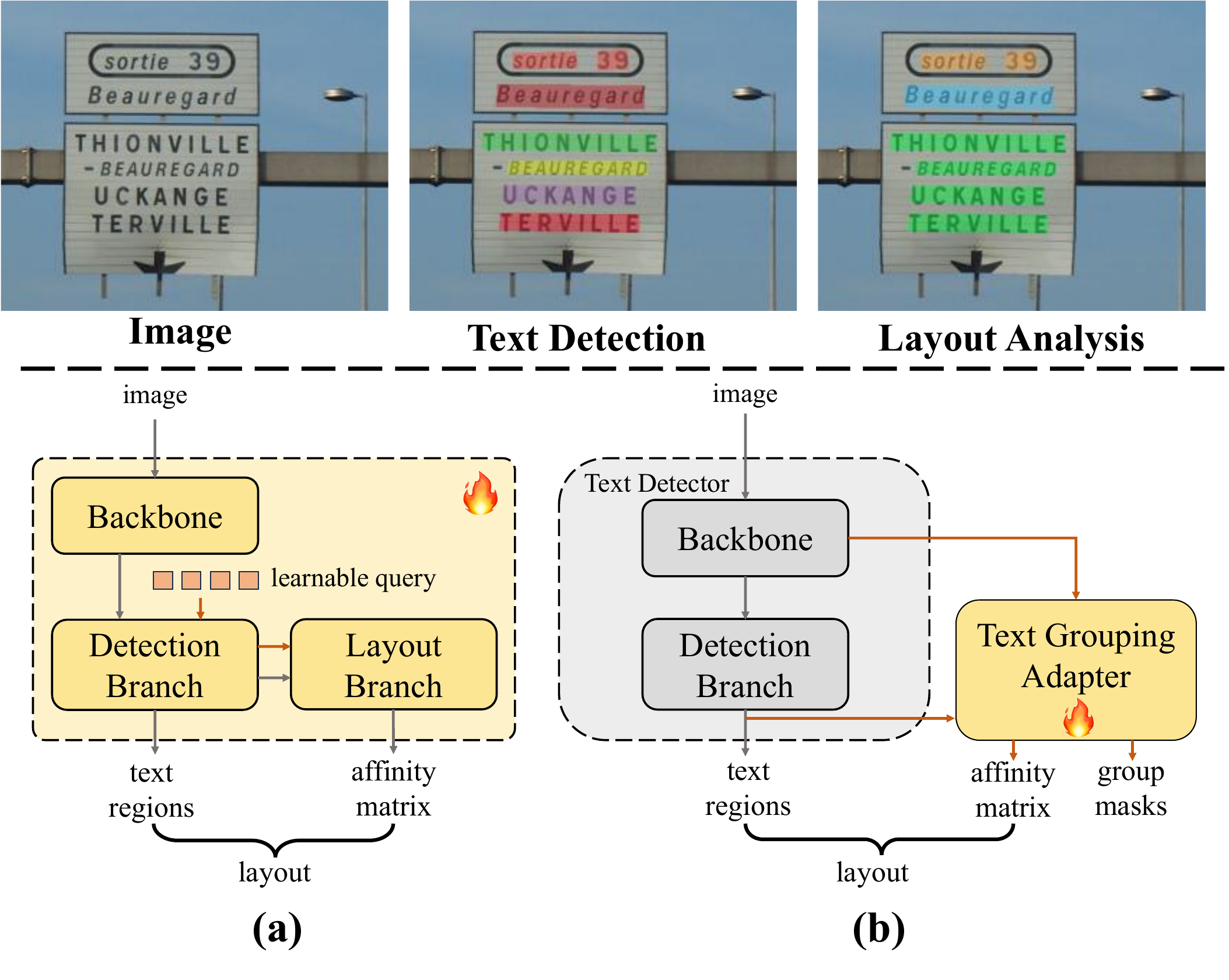}
    \caption{\textbf{Top}: Visualization of the scene text detection and layout analysis tasks. The mask with the same color denotes detected as a group. \textbf{Bottom}: Comparison between \textbf{(a)} the previous work Unified Detector~\cite{long2022towards} and \textbf{(b)} proposed TGA. 
    TGA also provides the flexibility of freezing or fine-tuning the pre-trained text detector.}
    \label{fig:intro}
    \vspace{-5mm}
\end{figure}

With the rise of deep learning, text detection~\cite{zhang2016multi,long2018textsnake,liao2020dbnet,liao2022dbnetpp}, text 
recognition~\cite{bluche2017gated, wang2020decoupledrec, diaz2021rethinking, li2023trocr}, and end-to-end text spotting~\cite{liu2020abcnet,liu2021abcnetv2,ye2023deepsolo,ye2023deepsolopp} models have greatly improved the accuracy and efficiency of identifying text instances like words or text lines. 
To fully understand the text semantics
in various applications~\cite{bissacco2013photoocr,shi2017detecting,9551780,binmakhashen2019document, zhang2023responsible}, it is essential to determine how to organize these text instances into coherent semantic entities, \egno, determining which detected words constitute a line and which detected lines form a paragraph.
This problem, as visualized in the top of Figure~\ref{fig:intro}, named scene text layout analysis~\cite{long2022towards}, has not been advanced at the same pace as other scene text understanding tasks.

Previous layout analysis works~\cite{lee2019page,zhong2019publaynet,schreiber2017deepdesrt} adopt
separate semantic segmentation models to localize different high-level entities in scanned and digital document images, which only focus on limited scenarios instead of general-scope natural scenes and ignore low-level text instances like words and text lines. 
Recent design on a Unified Detector~\cite{long2022towards} is the first to address the layout analysis problem in this
field and propose a unified model to detect text instances and their layouts. 
Different from previous works that model layout as high-level text entity localization, Unified Detector considers it 
from the perspective of the affinities between low-level text instances. This 
enables a direct prediction of 
the affinities between text instances.
The masks of high-level text entities can be derived by connecting these low-level instance masks.
However, this Unified Detector requires a query-based network structure and only considers jointly training its text detection branch with the layout branch from scratch. This implies a lack of flexibility in the detection network structure. More importantly, it only 
benefits from the layout analysis dataset, given the significantly smaller size of the existing layout analysis dataset, \egno, 8,281 images in HierText~\cite{long2022towards} compared to 30,000 in the text detection dataset IC19-LSVT~\cite{sun2019ic19lsvt}, which obviously limits the potential of text detection.

Given these limitations, we pose the question:  \textit{can we enable already pre-trained text detectors with a new module to learn layout analysis?} 
Nonetheless, answering this question presents nontrivial challenges. 
The first challenge arises from the diversity of existing text detectors in terms of architecture and text region representations. Specifically, these text detectors employ a wide array of network structures, ranging from query-based transformers~\cite{ye2023deepsolo,zhang2022text,raisi2021transformer} to fully convolutional networks~\cite{zhang2016multi,liao2020dbnet,long2018textsnake} and dynamic convolutional networks~\cite{zhang2021knet}. Moreover, they also model the text region diversely, such as semantic mask~\cite{zhang2016multi,liao2020dbnet}, instance mask~\cite{long2022towards} and parameterized curve~\cite{liu2020abcnet,zhu2021fourier}. 
Another challenge is the lack of global features in pre-trained text detectors since it might overly focus on local features for text instance detection when pre-training. 

To address these challenges, we introduce Text Grouping Adapter (TGA), a novel module that adapts diverse pre-trained text detectors to learn layout analysis. 
Not only does it provide flexibility and compatibility for network structures, it also empowers the model to inherit a robust text detection capability from pre-training on large-scale text datasets.
Specifically, the TGA takes text regions and image features as input, and outputs affinities between regions to represent layouts following the approach of the Unified Detector. 
The TGA comprises two key components: Text Instance Feature Assembling (TIFA) and Group Mask Prediction (GMP). 
By seamlessly converting the text regions into text instance masks and embedding image feature into pixel embedding map, our TIFA ensures TGA can obtain text instance representations from the two inputs. 
It becomes the cornerstone of TGA's compatibility with various text detection architectures. 
GMP is designed as an additional task during training, to boost the learning of cohesive features for text instances within their respective groups, which helps the instances to understand the group region and aggregate global features necessary for layout analysis.
By doing so, TGA effectively bridges the gap between pre-trained text detectors and scene text layout analysis.

Our extensive experiments reveal that even when freezing the pre-trained models, 
the integration of our TGA with various pre-trained text detectors and text spotters can significantly enhance performance on the layout analysis dataset.
Moreover, it also allows for the inheritance of a generalized text detection ability from pre-training. 
Remarkably, when full parameter fine-tuning is applied, our TGA can further improve layout analysis performance. 
Varied ablation studies on the components of TGA are further conducted, demonstrating their substantial contribution to the improvement of layout analysis performance.
We believe such a Text Grouping Adapter will accelerate the development and application of layout analysis via leveraging broader pre-trained models and datasets.

\vspace{-3mm}
\section{Related Works}

\subsection{Text Detection and Spotting}

As an important and active topic in the computer vision field with numerous practical applications, text detection has been studied extensively.
However, to date, these works exhibit a high degree of variance in terms of text region representations and network structures.
For text representations, a series of works~\cite{zhang2016multi,liao2020dbnet, liao2022dbnetpp,wang2019shape,long2018textsnake} model the text regions as semantic masks. 
As alternative approaches, \cite{zhu2021fourier,liu2020abcnet,ye2023deepsolo,zhang2022text,raisi2021transformer} propose to leverage parameterized curves such as Bezier curves and Fourier contours to adaptively fit highly-curved text regions.
With regards to network structures, some works~\cite{zhang2016multi,liao2020dbnet, liao2022dbnetpp,wang2019shape,long2018textsnake} use fully convolutional networks to predict semantic masks, while others incorporate diverse approaches like query-based transformers~\cite{ye2023deepsolo,zhang2022text,raisi2021transformer} and region-based convolutional networks~\cite{liu2020abcnet,qin2019maskalign}.


\vspace{-1mm}
\subsection{Text Layout Analysis}
\vspace{-1mm}

While text detection has been extensively developed and studied, layout analysis remains in its nascent stages, particularly concerning scene text, due to the complex and challenging task of distinguishing relationships between text instances.
Some works~\cite{schreiber2017deepdesrt,zhong2019publaynet,lee2019page} propose to analyze the layout of document images, where the task is defined as to localize semantically coherent text blocks. 
These works only focus on document images and neglect word or line-level text instances.
Long et al.~\cite{long2022towards} is the first work to study layout analysis in the scene text field and train a unified model to detect text instances and recognize their relationships as layout.
Given the fact that layout analysis annotations are expensive and rare, training a unified scene text layout analysis model only based on the layout analysis dataset poses limitations in learning text instance detection capabilities.
Our work aims to harness the advancements made in scene text detection to enhance layout analysis.

\vspace{-2mm}
\subsection{Adapter}
\vspace{-1mm}
Adapters have garnered extensive utilization in the Natural Language Processing (NLP) field as an efficient tool to enable the adaptation of pre-trained models to downstream NLP tasks.
In the field of computer vision, earlier works apply adapters with incremental learning~\cite{rosenfeld2018incremental} and domain adaptation~\cite{rebuffi2017learning, rebuffi2018efficient}. Recent works focus on leveraging adapters to transfer pre-trained vision transformers into downstream tasks, including dense prediction~\cite{chen2022vision} and vision-language domains~\cite{li2023blip,gao2023llama,zhu2023minigpt}. Similar to the previous adapter works on other domains, our TGA transfers the pre-trained text detector into the layout analysis task to inherit the knowledge learned from pre-training.

\vspace{-2mm}
\section{Methods}
\vspace{-1mm}
\begin{figure*}
    \centering
    \includegraphics[width=0.9\linewidth]{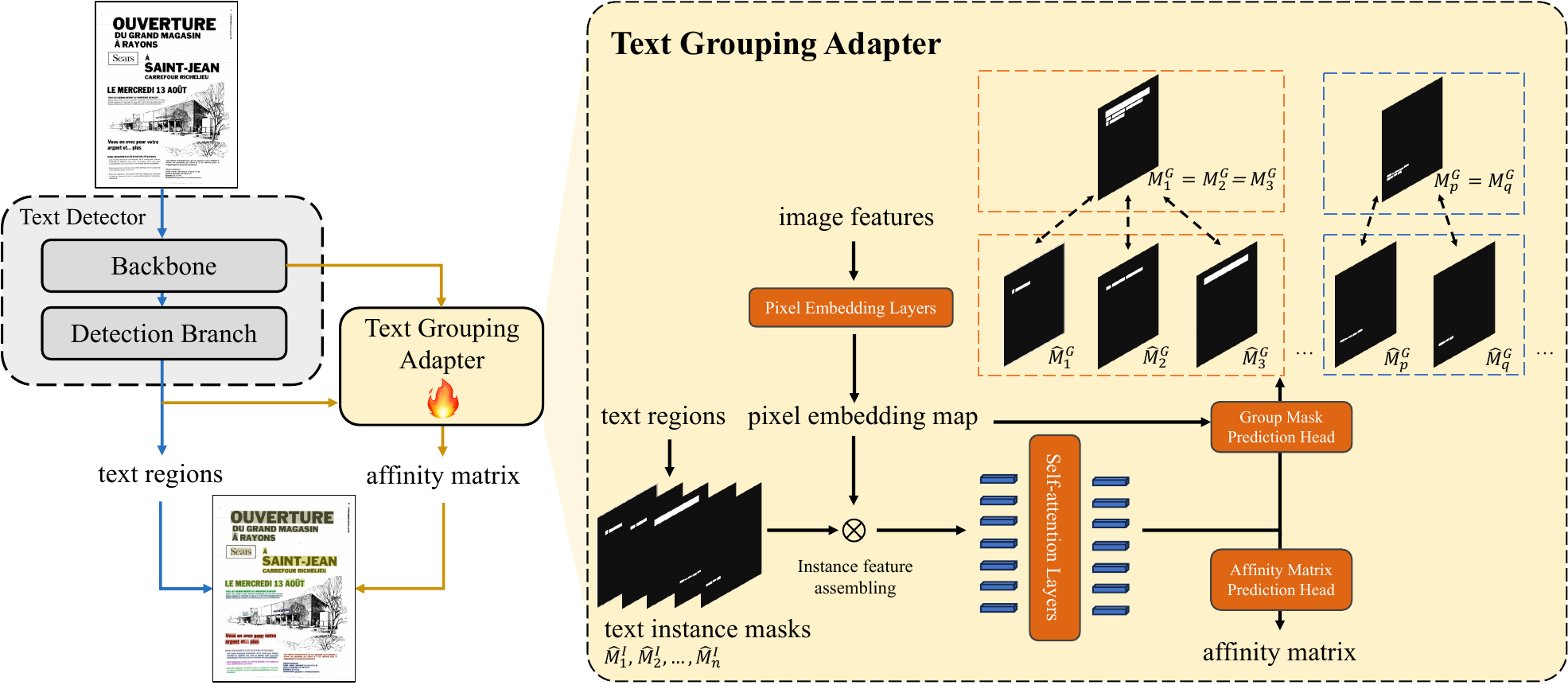}
    \caption{Overview of proposed Text Grouping Adapter. The dashed boxes denote the matched group and instance.
    The text detector can be frozen or fine-tuned together with TGA when training. 
    $\hat{\mathbf{M}}^{I}_{i}$ is the predicted instance mask of $I_i$. $\hat{\mathbf{M}}^{G}_{i}$ and $\mathbf{M}^{G}_{i}$ are predicted group mask of $I_i$ and assigned ground-truth one of $I_i$. To illustrate, a same group mask is duplicated as $\mathbf{M}^{G}_{p}$ and $\mathbf{M}^{G}_{q}$ and assigned to $I_p$ and $I_q$.
    }
    \vspace{-5mm}
    \label{fig:overview}
\end{figure*}

Figure~\ref{fig:overview} illustrates the pipeline of our proposed Text Grouping Adaptor (TGA), including two key components: Text Instance Feature Assembling (TIFA) and Group Mask Prediction (GMP).
TGA takes image features and detected text regions as inputs, producing affinities between text regions. 
Within the TIFA component, text instance features are assembled, and GMP encourages these instance features to learn cohesive group features during training.
Then, the learned text instance features are used to predict the final affinity matrix.
We now discuss these components in detail.

\vspace{-1mm}
\subsection{Text Instance Feature Assembling}
\vspace{-1mm}

Text Instance Feature Assembling (TIFA) is designed to process diverse representations of text regions and produce text instance features as output.
This requires a unified text region representation and a spatially correspondent approach to assemble instance features.
Inspired by instance segmentation methods~\cite{cheng2021maskformer,zhang2021knet}, we employ a pixel embedding map and instance masks to assemble text instance features. We sequentially introduce this module as follows.

\vspace{-5mm}
\paragraph{Unified text region representation.} Various text detectors generate different representations of text regions, e.g., semantic masks, instance mask, and kinds of parameterized curves. 
For the compatibility of TGA and instance-level understanding, we select the text instance mask as our representation, which can be converted seamlessly from other representations.
Specifically, for semantic masks, we employ binarization on each mask and utilize the algorithm~\cite{suzuki1985findcontour} to identify text instances within them.
In the case of parameterized curves, we transform each curve instance into a close polygon, then draw and fill the polygon to create the corresponding instance mask.
We define 
$\mathbf{I} = \{I_i\}_{i=1}^{N}$ 
as the extracted text instance set, including $N$ instances. 
Empty masks are padded when less than $N$ instances are found in the output of text detectors.
By this process, we unify disparate text region representations as text instance masks $\hat{\mathbf{M}}^{I}$ for later feature assembling.

\vspace{-4mm}
\paragraph{Pixel embedding map.}

%

Grouping text instances necessitates both low-level features to discriminate small-size texts and high-level features to provide sufficient contextual information.
For this, we derive a pixel embedding map from multi-scale features  $\{\mathbf{X}_2, \mathbf{X}_3, \mathbf{X}_4, \mathbf{X}_5\}$, of the text detector's backbone. 
The sizes of $\{\mathbf{X}_2, \mathbf{X}_3, \mathbf{X}_4, \mathbf{X}_5\}$ are $\{\frac{1}{4}, \frac{1}{8}, \frac{1}{16}, \frac{1}{32}\}$ of the input image size. Through a series of convolutional layers and upsampling, each $\mathbf{X}_i$ is transformed into a corresponding $\mathbf{P}_i$, all standardized to $\frac{1}{8}$ of the input image size. The final pixel embedding map $\mathbf{P}$ is obtained by summing up $\{\mathbf{P}_l | i=1, ..., n\}$  and then feeding the result to another convolution to refine the pixel features.
With a simple fully convolutional network, we embed the multi-scale features into the pixel embedding map, ensuring the retention of distinct text features while capturing the necessary contextual information for layout analysis. We refer this simple fully convolutional network as Pixel Embedding Layers, shown in the Figure~\ref{fig:overview}.



\vspace{-4mm}
\paragraph{Feature assembling.}

As the converted instance mask prediction $\hat{\mathbf{M}}^{I}$ determines if a pixel of the image belongs to the text instance, we can assemble the instance feature by integrating the pixel embedding map $\mathbf{P}$ and the text instance mask $\hat{\mathbf{M}}^{I}$. 
This is accomplished through a multiplication operation between the two, which is equivalent to extracting the $D$-dimensional feature for each point corresponding to the pixel embedding map and sum pooling spatially.


\begin{footnotesize}
\begin{equation}
    \mathbf{F}^{I} = \hat{\mathbf{M}}^{I} \cdot \mathbf{P}^{\top}, 
\end{equation}
\end{footnotesize}

\noindent where $\hat{\mathbf{M}}^{I} \in \mathbb{R}^{N \times H \times W}$ refers to $N$ text instance masks converted from the prediction of the text detector, and $\mathbf{F}^{I} \in \mathbb{R}^{N \times D}$ are $N$ text instance features.

\vspace{-1mm}
\subsection{Group Mask Prediction}
\vspace{-1mm}

The text instance features assembled from individual instance masks in TIFA exclusively focus on their respective instance regions, overlooking the contextual information.
Hence, we introduce Group Mask Prediction (GMP) subsequent to TIFA. 
GMP encourages text instances to learn cohesive features of their corresponding group, thereby realizing the implicit clustering, which is crucial for the accurate prediction of the final affinity matrix.

To facilitate interaction among text instances and extract contextual information in group mask prediction, we first incorporate self-attention layers into the process. We fed text instance features into the self-attention layers: $\hat{\mathbf{F}}^{I} = SA(\mathbf{F}^{I}, \mathbf{F}^{I})$, where $\hat{\mathbf{F}}^{I} \in \mathbb{R}^{N \times D}$ denotes the updated text instance features and $SA(\cdot)$ denotes self-attention layers.


Then, we set the group mask with a one-to-many assignment as the learning target. 
Specifically, for each text group entity, \egno, text paragraph, we assign the group's mask to all instances that belong to that group using Hungarian matching. 
To achieve a more accurate assignment, we avoid direct matching of the detected instance masks with the ground-truth group masks.
Instead, we first match the detected instance masks with the ground-truth instance masks.
Subsequently, we transform the matched ground-truth instance masks into the corresponding ground-truth group masks according to the provided annotations.
The instance-level matching result are denoted as $\hat{\sigma}$:

\begin{footnotesize}
\begin{equation}
    \hat{\sigma} = \underset{\sigma \in \mathfrak{S}_{N}}{\arg \min} \sum_{i}^{N} \mathcal{L}_{match}(\mathbf{M}^{I}_{i}, {\hat{\mathbf{M}}^{I}_{\sigma(i)}}),
\end{equation}
\end{footnotesize}

\noindent where we match for a permutation of $N$
text instances $\sigma \in \mathfrak{S}_{N}$ with the lowest cost. 
Under certain $\sigma$,  $\mathbf{M}^{I}_{i}$ is the ground-truth text instance mask, while $\hat{\mathbf{M}}^{I}_{\sigma(i)}$ is its matched predicted value. 
$\mathcal{L}_{match}$ is the pair-wise matching cost, which is 
 designed to be consistent with the original loss of text detectors, detailed in our supplementary material.

We then obtain the predicted group mask $\hat{\mathbf{M}}^{G}$ via a dot-product between the updated text instance features and the pixel embedding map: 
\begin{footnotesize}
\begin{equation}
    \hat{\mathbf{M}}^{G} = \hat{\mathbf{F}}^{I} \otimes \mathbf{P},
\end{equation}
\end{footnotesize}

\noindent where $\hat{\mathbf{M}}^{G} \in \mathbb{R}^{N \times H \times W}$ are the $N$ predicted text group masks from the $N$ text instances. 
Subsequently, we compute the Dice Loss between predicted text group masks and matched ground-truth group masks:

\begin{footnotesize}
\begin{equation}
    \begin{aligned}
        \mathcal{L}_{dice} = 1 - \frac{2 \sum_{i}^{N} \mathbf{M}^{G}_{i} \cdot \hat{\mathbf{M}}^{G}_{\hat{\sigma}(i)}}{\sum_{i}^{N} \left \lVert \mathbf{M}^{G}_{i} \right \rVert_{F}^2 + \sum_{i}^{N} \left \lVert \hat{\mathbf{M}}^{G}_{\hat{\sigma}(i)} \right \rVert_{F}^2},
    \end{aligned}
\end{equation}
\end{footnotesize}

\noindent where the predicted instances without matched group-truth ones are excluded from the loss calculation.

This approach transforms DETR-style~\cite{carion2020detr} one-to-one assignment to a one-to-many assignment by replacing matched instance masks with corresponding group masks, resulting in the same group mask being assigned to all text instances belonging to the group.
It also differs from the one-to-many assignment concept referred to in other DETR variants~\cite{jia2023detrs,chen2023group}, which assign single instance supervision into many instance predictions for faster convergence rather than implicitly clustering instances for group entities.

\begin{figure}
    \centering
    \includegraphics[width=0.8\linewidth]{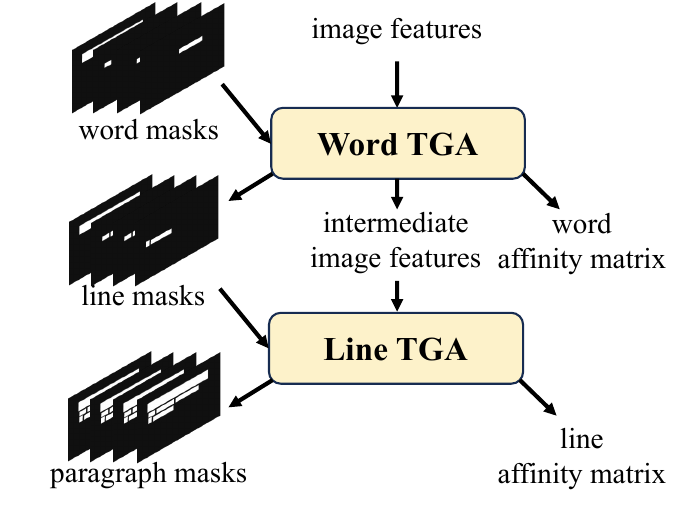}
    \caption{Details of Cascade TGA. Intermediate image features denotes the multi-scale features before summing in Word TGA.}
    \label{fig:cascade}
    \vspace{-6mm}
\end{figure}

\vspace{-1mm}
\subsection{Affinity Matrix Prediction}
\vspace{-1mm}

Following Unified Detector, we predict an affinity matrix between text instances to represent the layout. By multiplying the updated text instance features with their transpose, we can obtain the predicted affinity matrix $ \hat{\mathbf{A}} = \hat{\mathbf{F}}^{I} \cdot {(\hat{\mathbf{F}}^{I})}^{\top}$,
where the element $\hat{\mathbf{A}}_{i, j}$ denotes the predicted affinity score between $I_i$ and $I_j$, \ieno, the possibility of the instance pair belonging to the same group.
We utilize the previously obtained instance matching result $\hat{\sigma}$ and annotations to construct ground truth binary affinity matrix $\mathbf{A} \in \mathbb{R}^{N \times N}$ and the binary instance loss weight $\mathbf{C} \in \mathbb{R}^{N \times N}$.
The element $\mathbf{A}_{i, j} \in \{0, 1\}$ is set to 1 if the pair, $I_i$ and $I_j$, belongs to the same text group and 0 otherwise.
$\mathbf{C}_{i, j}$ is used to exclude the instances without matched ground truth under $\hat{\sigma}$.

Finally, we calculate the binary cross-entropy loss between the predicted and ground-truth affinity matrix as the loss function for text grouping:

\begin{footnotesize}
\begin{equation}
    \begin{aligned}
        \mathcal{L}_{group} = \sum_{i}^{N} \sum_{j}^{N} \mathbf{C}_{i, j} [
        & \mathbf{A}_{i, j} \cdot \log (\hat{\mathbf{A}}_{i, j})\ + \\[-2ex]
        & (1 - \mathbf{A}_{i, j}) \cdot \log (1 - \hat{\mathbf{A}}_{i, j})].
    \end{aligned}
\end{equation}
\end{footnotesize}

\noindent during the inference phase, we set a group threshold $t$ on the affinity matrix $\mathbf{A}$ to determine whether pairs of texts are related. We then use the union-intersection approach to obtain the final text grouping.

\vspace{-1mm}
\subsection{Loss Function}
\vspace{-1mm}

The TGA's overall loss function is calculated as the weighted sum of the group mask prediction, affinity matrix, and optional original text detection loss functions:

\begin{footnotesize}
\begin{equation}
    \mathcal{L}_{TGA} = \alpha_1 \mathcal{L}_{dice} + \alpha_2 \mathcal{L}_{group} + \alpha_3 \mathcal{L}_{det},
\end{equation}
\end{footnotesize}

\noindent where $\mathcal{L}_{det}$ is the sum of original text detection loss functions. $\alpha_1$, $\alpha_2$ and $\alpha_3$ represent the weights of GMP, AMP, and original detection loss functions respectively. 






\vspace{-1mm}
\subsection{Cascade TGA for Word-based Text Detector}
\vspace{-1mm}

Different from directly grouping detected lines into paragraphs, 
grouping detected words into paragraphs
is actually a hierarchical grouping task. 
It's more challenging as the words in a paragraph might exhibit greater diversity and be more spatially distant from each other.
Thanks to the compatibility of the proposed TGA, we can simply cascade more than one TGA to address this problem, referred as Cascade TGA.
As shown in Figure~\ref{fig:cascade}, we cascade two TGAs, named Word TGA and Line TGA.
Detected text word regions and image features are first fed into Word TGA to predict line masks.
The predicted line masks are subsequently fed into Line TGA to predict paragraph masks and the affinity matrix. 
This design complements the mid-level supervision and showcases the compatibility of TGA.
\vspace{-5mm}
\section{Experiments}

\begin{table*}[ht]
\centering
\resizebox{\textwidth}{!}{%
\begin{tabular}{@{}lcccccccccc@{}}
\toprule
\multicolumn{1}{c}{\multirow{3}{*}{Models}} & \multirow{3}{*}{\begin{tabular}[c]{@{}c@{}}Frozen\\ Text Detector\end{tabular}} & \multirow{3}{*}{\begin{tabular}[c]{@{}c@{}}Trainable Params / \\ Total Params\end{tabular}} & \multicolumn{4}{c}{Val} & \multicolumn{4}{c}{Test} \\ \cmidrule(l){4-11} 
\multicolumn{1}{c}{} &  &  & \multicolumn{2}{c}{Instance} & \multicolumn{2}{c}{Paragraph} & \multicolumn{2}{c}{Instance} & \multicolumn{2}{c}{Paragraph} \\ \cmidrule(l){4-11} 
\multicolumn{1}{c}{} &  &  & F & PQ & F & PQ & F & PQ & F & PQ \\ \midrule
\multicolumn{11}{l}{\textit{Word-based}} \\ \midrule
Unified Detector & \ding{56} & 88.2M / 88.2M & \multicolumn{1}{l}{\textbf{78.70}} & \multicolumn{1}{l}{\textbf{62.84}} & \multicolumn{1}{l}{50.11} & \multicolumn{1}{l}{39.26} & \multicolumn{1}{l}{\textbf{80.03}} & \multicolumn{1}{l}{\textbf{63.89}} & \multicolumn{1}{l}{51.54} & 40.33 \\
TGA + Deepsolo-ViTAE-S & \ding{52} & 8.7M / 44.8M & \multicolumn{1}{l}{63.45} & \multicolumn{1}{l}{47.20} & \multicolumn{1}{l}{55.97} & \multicolumn{1}{l}{40.32} & \multicolumn{1}{l}{64.34} & \multicolumn{1}{l}{47.94} & \multicolumn{1}{l}{55.59} & 40.18 \\
TGA + DBNetPP-R50 & \ding{52} & 8.0M / 34.1M & \multicolumn{1}{l}{76.91} & \multicolumn{1}{l}{58.70} & \multicolumn{1}{l}{\textbf{56.23}} & \multicolumn{1}{l}{\textbf{42.23}} & \multicolumn{1}{l}{78.07} & \multicolumn{1}{l}{59.60} & \multicolumn{1}{l}{\textbf{56.32}} & \textbf{42.23} \\ \midrule
\multicolumn{11}{l}{\textit{Line-based}} \\ \midrule
GCP API & - & - & - & - & - & - & - & 56.17 & - & 46.33 \\
Max-DeepLab-Cluster & - & - & - & - & - & - & - & 62.23 & - & 52.52 \\
Unified Detector & \ding{56} & 88.2M / 88.2M & 78.44 & 61.04 & 67.76 & 52.84 & 79.91 & 62.23 & 68.58 & 53.60 \\
TGA + KNet-R50 & \ding{56} & 43.4M / 43.4M & 77.04 & 59.08 & 71.90 & 55.27 & 77.88 & 60.03 & 71.37 & 55.04 \\
TGA + MaskDINO-R50 & \ding{56} & 52.2M / 52.2M & 76.41 & 58.91 & 75.46 & 58.28 & 77.43 & 59.89 & 75.41 & 58.33 \\
TGA + KNet-R50 & \ding{52} & 5.9M / 43.4M & 76.66 & 58.68 & 71.31 & 54.58 & 77.37 & 59.54 & 70.34 & 54.05 \\
TGA + MaskDINO-R50 & \ding{52} & 5.9M / 52.2M & 84.10 & 64.51 & 74.67 & 57.62 & 84.92 & 65.35 & 74.22 & 57.47 \\
TGA + KNet-Swin-B & \ding{52} & 5.9M / 106.2M & 79.10 & 60.64 & 73.58 & 56.41 & 79.67 & 61.40 & 72.49 & 55.90 \\
TGA + MaskDINO-Swin-B & \ding{52} & 7.1M / 123.0M & \textbf{85.84} & \textbf{66.50} & 75.45 & 58.72 & \textbf{86.65} & \textbf{67.27} & 75.11 & 58.65 \\
TGA$^{\dag}$ + MaskDINO-R50 & \ding{52} & 12.0M / 58.2M & 85.00 & 65.60 & \textbf{78.00} & \textbf{60.28} & 86.18 & 66.63 & \textbf{77.61} & \textbf{60.05} \\ \bottomrule
\end{tabular}%
}
\caption{Results of word and line detection on the HierText Dataset for all models. 
\textit{Word-based} refers to detects and groups word regions as paragraphs while \textit{Line-based} does the same for text line regions. The fine-tune strategy and trainable parameter size is inapplicable for API or non-unified model.
Deepsolo-ViTAE-S means the Deepsolo with ViTAE-S~\cite{xu2021vitae} as its backbone.
KNet-R50 means the KNet with ResNet-50~\cite{he2016resnet} as its backbone. 
Swin-B denotes Swin-Base~\cite{liu2021swin}. Other models' naming follows the same approach.
TGA$^{\dag}$ means an enhanced TGA by adding more layers in Pixel Embedding Layers for more trainable parameters while other settings keep the same.}
\label{tab:main-results}
\vspace{-6mm}
\end{table*}

In this section, we set up comprehensive experiments and analysis on TGA. 
We incorporate TGA into diverse pre-trained text detectors and compare their performance with a series of competitive baseline methods. 
To explore the possibility of parameter-effectively adapting the text detectors, we also compare the TGA's performance on diverse datasets under the condition that the parameters of the text detectors are either frozen or not.
Finally, we validate the effectiveness of TGA's components by extensive ablation studies and visualizations.

\vspace{-1mm}
\subsection{Experiment Settings}
\vspace{-1mm}

\paragraph{Baselines.}
Since the layout analysis can build on word instances and text line instances, we compare methods separately on word-based layout analysis and line-based layout analysis.
Since there is little literature specially studying word-based layout analysis, we solely draw on the Unified Detector~\cite{long2022towards} as our word-based baseline.
As for line-based methods, we adapt baselines from \cite{long2022towards}, including Google Cloud OCR API~\cite{google_ocr}, Max-DeepLab-Cluster~\cite{wang2021maxdeeplab,long2022towards} and the Unified Detector~\cite{long2022towards}. Google OCR (GCP) API is a public commercial OCR engine. 
Max-DeepLab-Cluster builds two separate Max-DeepLab models to detect text lines and paragraphs and reassign the affinities between instances by post-processing. The Unified Detector stands out as the first unified model to detect text instances and layout simultaneously, achieving the previous state-of-the-art performance. Note that in the following comparison, we refer to the Unified Detector with 384 queries for its best performance.

\vspace{-4mm}
\paragraph{Pre-trained text detectors.} We applied TGA to the following four models pre-trained for text detection with diverse network structures and text region representations: DBNetpp~\cite{liao2020dbnet}, Deepsolo~\cite{ye2023deepsolo}, KNet~\cite{zhang2021knet}, and MaskDINO~\cite{li2023maskdino}.
In terms of the network architecture, DBNetpp utilizes a fully convolutional network, while MaskDINO and DeepSolo employ query-based transformers. 
KNet introduces dynamic kernels with the convolutional network.
Regarding the text region representations, DBNetpp produces semantic masks.
KNet and MaskDINO model the text region as text instance masks. 
DeepSolo stands out as the state-of-the-art text spotting model, outputting Bezier Curve control points.
Besides the differences resulting from their methods, in our experiments, DBNetpp is pre-trained as word-level text detectors, while DeepSolo serves as the word-level text spotting model. KNet and MaskDINO are pre-trained as line-level text detectors. 

\vspace{-4mm}
\paragraph{Datasets and metrics.} 
HierText~\cite{long2022towards} is the scene text layout analysis dataset for our training and evaluation, which is well annotated with not only paragraph masks but also text line and word masks.
It consists of 8,281, 1,724, and 1,634 images in training, validation, and test set, respectively.
Following~\cite{long2022towards}, we use Precision (P), Recall (R), F1 score (F), and Panoptic Quality (PQ) as our metrics for layout analysis and text detection on the HierText Dataset, where PQ is computed as the product of the F1 score and the average Intersection over Union of all True Positive pairs.

For text detection, CTW1500~\cite{yuliang2017detecting}, MSRA-TD500~\cite{yao2012detecting}, IC19-LSVT~\cite{sun2019ic19lsvt} and HierText, totally 39,581 images, are the datasets used for the pre-training of line-level text detector. 
As for word-level text detection, we proceed with the continual pre-training of the DBNetpp model and directly draw the off-the-shelf parameters from the DeepSolo model.
We use Average Precision (AP) and the Harmonic Mean (H-mean) as the text detection metrics.
We report more details of datasets, metrics and pre-training in our supplementary material.
\vspace{-4mm}
\paragraph{Hyperparameter setting.} Hyperparameters of TGA network structure are configured as follows: the feature dimension, $D$, is set to 256, the number of self-attention layers is 3, the number of attention heads is 4, and the dimension of the hidden layers is 512. 
During the inference phase, we set the group threshold $t$ as 0.8 for all these models.
As for training hyperparameters, we mainly refer to the original text detector training hyperparameters, which are provided in our supplementary material.

\vspace{-1mm}
\subsection{Main Comparison} 
\label{exp:main-results}
\vspace{-1mm}

\begin{table*}[ht]
\centering
\resizebox{0.9\textwidth}{!}{%
\begin{tabular}{@{}lcccccccc@{}}
\toprule
\multicolumn{1}{c}{\multirow{2}{*}{Models}} & \multirow{2}{*}{\begin{tabular}[c]{@{}c@{}}Frozen \\ Text Detector\end{tabular}} & \multirow{2}{*}{\begin{tabular}[c]{@{}c@{}}Trainable \\ Params\end{tabular}} & \multirow{2}{*}{\begin{tabular}[c]{@{}c@{}}Training \\ Time\end{tabular}} & CTW1500 & MSRA-TD500 & IC19-LVST & \multicolumn{2}{c}{HierText Val} \\ \cmidrule(l){5-9} 
\multicolumn{1}{c}{} &  &  &  & Line AP & Line AP & Line H-mean & Line PQ & Paragraph PQ \\ \midrule
\multirow{2}{*}{TGA + KNet-R50} & \ding{56} & 43.4M & 11.5h & 43.12 & 47.54 & 69.89 & \textbf{60.59} & \textbf{51.85} \\
 & \ding{52} & 5.9M & 11.0h & \textbf{56.37} & \textbf{57.38} & \textbf{76.01} & 57.80 & 49.83 \\ \midrule
\multirow{2}{*}{TGA + KNet-Swin-B} & \ding{56} & 106.2M & 20.2h & 44.00 & 57.25 & 76.90 & \textbf{62.97} & \textbf{54.17} \\
 & \ding{52} & 5.9M & 13.2h & \textbf{58.28} & \textbf{59.81} & \textbf{78.69} & 59.75 & 50.77 \\ \midrule
\multirow{2}{*}{TGA + MaskDINO-R50} & \ding{56} & 52.2M & 31.8h & 41.46 & 51.80 & 59.19 & 59.88 & \textbf{59.19} \\
 & \ding{52} & 5.9M & 23.8h & \textbf{61.18} & \textbf{59.54} & \textbf{82.62} & \textbf{65.83} & 54.95 \\ \midrule
\multirow{2}{*}{TGA + MaskDINO-Swin-B} & \ding{56} & 123.0M & \multicolumn{6}{c}{Out of Memory} \\
 & \ding{52} & 7.1M & 27.3h & \textbf{61.25} & \textbf{59.58} & \textbf{82.63} & \textbf{66.37} & \textbf{58.56} \\ \bottomrule
\end{tabular}%
}
\caption{Comparison between different fine-tuning strategies: frozen text detector and full fine-tuning. Out of Memory denotes this error occurs when fully fine-tuning the TGA + MaskDINO-SWin-B on a 8 * V100-32G workstation with minimal batch size.
The values on HierText Val set slightly differ from ones reported in Table~\ref{tab:main-results} for slightly different annotations, detailed in supplementary material.}
\vspace{-3mm}
\label{tab:tuning-comparison}
\end{table*}

In this main comparison, we emphasize Paragraph F and PQ as our main metrics since they represent the performance of layout analysis. 
Instance F and PQ also are considered as better instance detection helps layout analysis.
As shown in Table~\ref{tab:main-results}, we incorporate TGA into various pre-trained text detectors, achieving impressive results on both word-based and line-based layout analysis.
On word-based comparison with the Unified Detector, with much smaller trainable parameter size and total parameter size (around 10\% trainable parameters and one-half total parameters), both frozen DBNetpp and DeepSolo combined with TGA achieve on-par, even superior performance on Paragraph PQ.
It's also noteworthy that, due to the limitation of the text detector's parameter size, our layout analysis results are produced given this worse instance-level performance, which further demonstrates our advantages in grouping semantic entities.

On line-based layout analysis, while applying TGA into different text detectors, we also involve the settings of whether or not to freeze the text detector.
As shown in the bottom section of Table~\ref{tab:main-results}, under all possible configurations, all our models consistently outperform all previous models on the Paragraph PQ of the test set to varying degrees.
Among them, even the lightest model, TGA with frozen KNet-R50 outperforms Unified Detector on Paragraph PQ by 0.8\% relatively. 
For the most powerful model, the TGA$^{\dag}$, adding more trainable parameters in Pixel Embedding Layers to get the enhanced pixel embedding map, achieves 60.05 PQ at the paragraph level with frozen MaskDINO-R-50. It is a 12.0\% relative improvement compared to the Unified Detector.
For our models with ResNet-50, we notice the fully fine-tuned models gain slightly better performance than the models with frozen text detectors on paragraph-level metrics, which is further studied in the following section.
We further show that TGA is robust to different annotation levels under the same model, detailed in supplementary material.
In summary, TGA is seamlessly compatible with various pre-trained text detectors and achieves superior layout analysis performance, even given the limitation of parameter size and worse detected text instances.



\begin{table}[ht]
\centering
\resizebox{0.80\columnwidth}{!}{%
\begin{tabular}{@{}ccccc@{}}
\toprule
\multirow{2}{*}{\begin{tabular}[c]{@{}c@{}}Frozen \\ Text Detector\end{tabular}} & \multicolumn{2}{c}{Components} & \multirow{2}{*}{\begin{tabular}[c]{@{}c@{}}Line \\ PQ\end{tabular}} & \multirow{2}{*}{\begin{tabular}[c]{@{}c@{}}Paragraph \\ PQ\end{tabular}} \\ \cmidrule(lr){2-3}
 & Pix. Emb. Map & GMP &  &  \\ \midrule
\multirow{3}{*}{\ding{56}} & \ding{56} & \ding{56} & \textbf{58.74} & 49.99 \\
 & \ding{56} & \ding{52} & 58.25 & 54.55 \\
 & \ding{52} & \ding{52} & 58.55 & \textbf{55.03} \\ \midrule
\multirow{4}{*}{\ding{52}} & \ding{56} & \ding{56} & \multirow{3}{*}{58.40} & 49.09 \\
 & \ding{56} & \ding{52} &  & 51.20 \\
 & \ding{52} & \ding{52} &  & \textbf{54.27} \\ \cmidrule(l){2-5} 
 & \ding{52} & $\bigcirc$ & 58.40 & 47.32 \\ \bottomrule
\end{tabular}%
}
\caption{Ablation studies of all modules in TGA. $\bigcirc$ means using line masks as the ground truth of GMP.}
\label{tab:ablation-modules}
\vspace{-4mm}
\end{table}

\begin{table}[ht]
\centering
\resizebox{6cm}{!}{%
\begin{tabular}{@{}cccccc@{}}
\toprule
\multirow{2}{*}{BCE Loss} & \multirow{2}{*}{Dice Loss} & \multicolumn{4}{c}{Paragraph} \\ \cmidrule(l){3-6} 
 &  & P & R & F & PQ \\ \midrule
\ding{52} & \ding{52} & 76.47 & 65.06 & 70.31 & 53.72 \\
\ding{52} & \ding{56} & 65.89 & 36.44 & 46.93 & 35.68 \\
\ding{56} & \ding{52} & \textbf{76.68} & \textbf{66.23} & \textbf{71.07} & \textbf{53.88} \\ \bottomrule
\end{tabular}%
}
\caption{Ablation studies of category for GMP loss function.}
\label{tab:ablation-loss-type}
\vspace{-6mm}
\end{table}

\vspace{-1mm}
\subsection{Comparison of Fine-Tuning Strategies}
\label{exp:tuning-comparison}
\vspace{-1mm}

We study the effect of different fine-tuning strategies when using TGA.
We fine-tune the same pre-trained text detectors on the HierText Dataset for layout analysis and evaluate them on both text detection datasets and layout analysis datasets.
We adopt two different fine-tune strategies:  (1) only fine-tune TGA's parameters and freeze the original text detector, referred to as \textit{frozen text detector}. (2) fine-tune all parameters, including original text detector and TGA, referred to as \textit{full fine-tuning}.
Results are shown in Table~\ref{tab:tuning-comparison}.
We can obverse full fine-tuning strategy consistently enhances the layout analysis performance of all models.
The most notable improvement is seen in MaskDINO-R50, where the full fine-tuning strategy surpasses the frozen text detector strategy by a significant 4.2 on Paragraph PQ.

On the other hand, the benefit of frozen text detector is also obvious in that it significantly saves the training time and GPU memory for the less trainable parameters, especially for the ones that have heavier backbones.
Frozen detector detector strategy saves 34.7\% training time for KNet-Swin-B and prevents out-of-memory error for MaskDINO-Swin-B fine-tuning.
Another advantage of frozen text detector is that it can keep the generalized detection ability obtained from pre-training and prevent overfit on the layout analysis dataset, especially given the fact that the current size of the layout analysis dataset is much smaller than broad text detection datasets.
As shown in the middle columns of Table~\ref{tab:tuning-comparison}, on CTW1500, MSRA-TD500 and LSVT detection dataset, the full fine-tuning models' performances rapidly drop on text detection metrics.
Representatively, full fine-tuning KNet-R50 relatively drops 23.5\%, 17.1\%, and 8.1\%, respectively on the three text detection datasets compared with frozen one.
It shows the trade-off choice of full fine-tuning or freezing text detector, where the former helps boost better layout analysis performance on the specific dataset, and the latter saves computing resources and benefits robustness on broader scenarios.

\vspace{-1mm}
\subsection{Ablation Study}
\label{exp:ablation}
\vspace{-1mm}

To validate the efficacy of the key components in our proposed TGA, we conducted a series of experiments where we respectively replace the Pixel embedding map with a single scale image feature and removed the Group Mask Prediction feature. 
These experiments were performed under two different fine-tuning strategies. 
We compared single TGA with Cascade TGA in the training process and analyze these components sequentially in the following paragraphs.


\vspace{-3mm}
\paragraph{Pixel embedding map.}
Pixel embedding map is designed to encode high-level features for layout contextual information while preserving low-level features for text instance details. 
As shown in Table~\ref{tab:ablation-modules}, the removal of the Pixel embedding map, regardless of whether the text detector is frozen or fully fine-tuned, negatively impacts the performance of layout analysis. 
This suggests that our proposed components are generally effective across different fine-tuning strategies.
The results also indicate that components play different roles under different fine-tuning strategies.
The removal of the pixel embedding map leads to a 3.03 drop in the Paragraph PQ under a frozen text detector, as compared to a 0.48 drop under full fine-tuning.
It indicates distinct scale features are needed between text detection pre-training and layout analysis fine-tuning. 
Hence, the removal of the pixel embedding map under a frozen text detector results in the loss of global contextual information, which significantly impairs the layout analysis performance.

\begin{figure}
    \centering
    \includegraphics[width=0.8\linewidth]{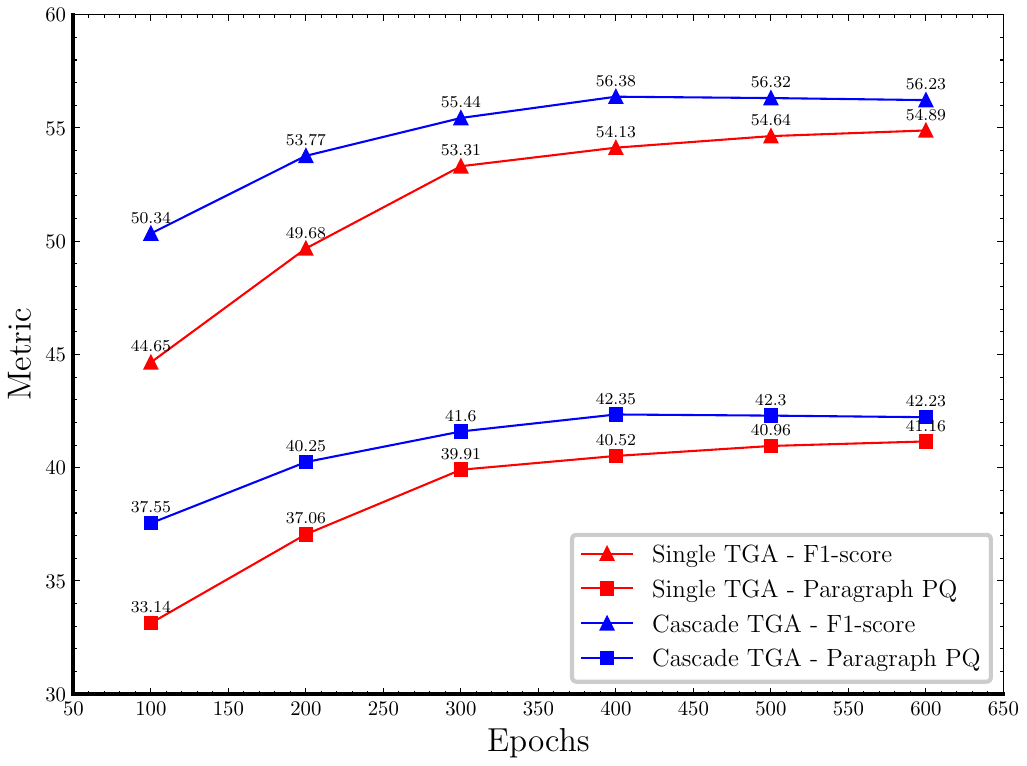}
    \caption{Comparison between different stages of single TGA and Cascade TGA training under frozen text detector strategy.}
    \vspace{-3mm}
    \label{fig:single-vs-cascade}
\end{figure}

\begin{figure}
    \centering
    \includegraphics[width=1.0\linewidth]{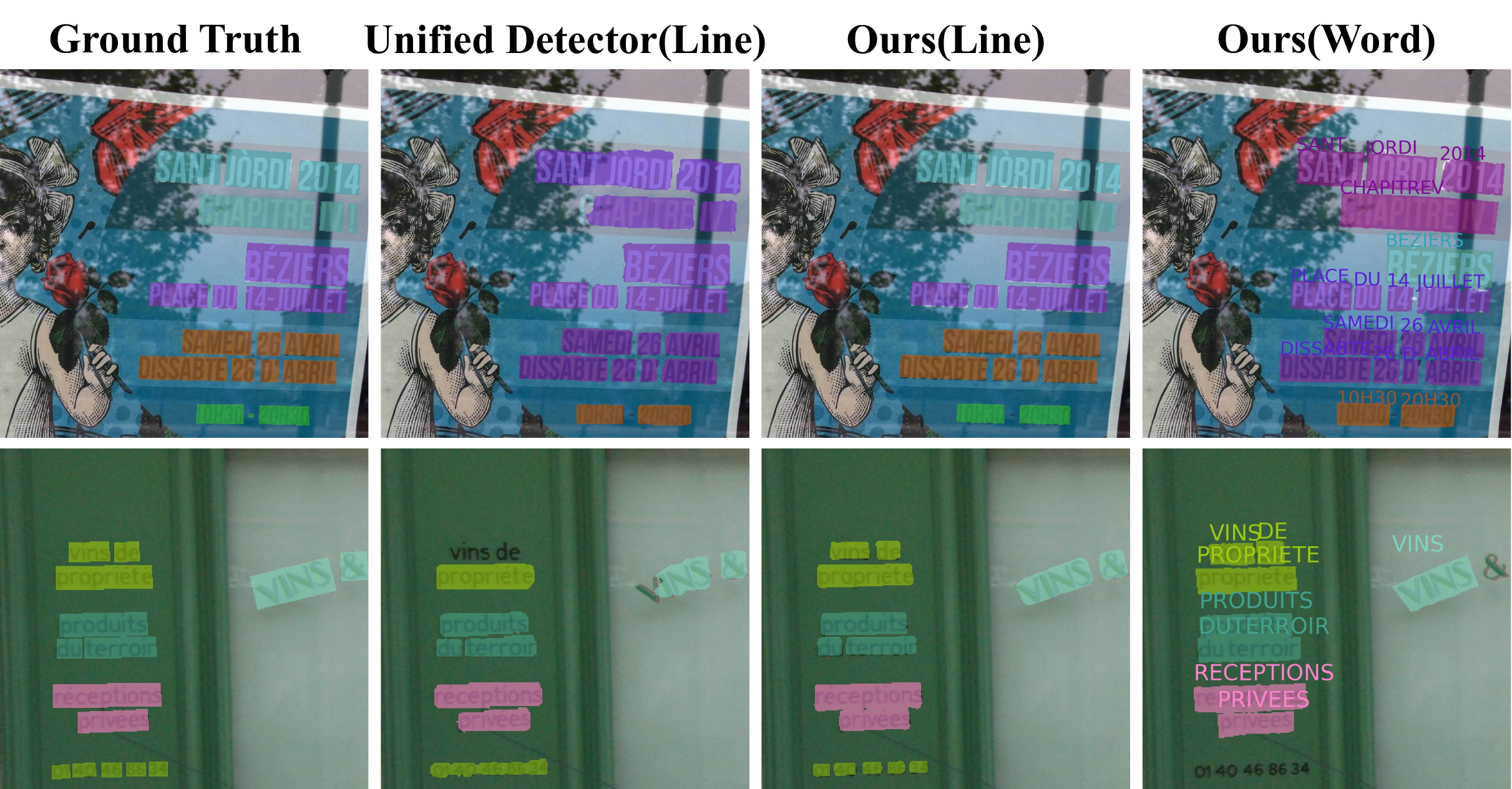}
    \caption{Visualization of results on the validation set of the HierText Dataset: from left to right, the sequence includes the ground truth, line-based Unified Detector, TGA + MaskDINO-Swin-B and TGA + DeepSolo-ViTAE-S. (Zoom in for the best view)}
    \vspace{-5mm}
    \label{fig:visualization}
\end{figure}

\vspace{-4mm}
\paragraph{Group Mask Prediction.} 
Group Mask Prediction (GMP) is a novel component in our Text Grouping Adapter, which leverages a one-to-many assignment to encourage text instances belonging to the same group to predict the same group mask. 
This process enables text instances to learn group-level features that encompass more contextual information.
As shown in Table~\ref{tab:ablation-modules}, models with GMP consistently outperform those without it.
Particularly under the full fine-tuning setting, solely add Group Mask Prediction gains 4.56 on Paragraph PQ.
When fine-tuned with the frozen text detector, the improvement is slightly less pronounced due to the limitation of detection-biased features.
With the assistance of the pixel embedding map, we still observe a substantial improvement of 5.18 on Paragraph PQ. 


\vspace{-5mm}
\paragraph{Group mask v.s. affinity matrix.}
The question arises as to \textit{why GMP helps the prediction of the affinity matrix given both group masks and affinity are derived from group annotations?}
To answer this question, we initially ablate the supervision used in GMP.
As shown in the last row of Table~\ref{tab:ablation-modules}, replacing the one-to-many assigned group mask with the one-to-one assigned line mask in the prediction causes the drop back to baseline performance levels. 
This confirms the advantages of GMP from the unique group masks and the one-to-many assignment, not merely from simple mask prediction. 
Unlike the affinity matrix, which depicts layout through pairwise relationships, the group mask represents layout through the collective representation of all instances within the group, thereby optimizing the inter-instance distances globally.
Our investigation into mask prediction loss combinations for GMP, shown in Table ~\ref{tab:ablation-loss-type}, demonstrates that relying solely on binary cross-entropy loss, computed pixel-wise, leads to a drastic decline in layout analysis performance.
Conversely, employing dice loss, which evaluates the holistic statistical resemblance between predicted group masks and their true counterparts, significantly elevates layout analysis outcomes. 
When using dice loss, which focus on the holistic statistic similarity between predicted group masks and ground-truth ones, it greatly boosts the performance of layout analysis.
These findings validate the GMP design's capacity to capture a more global and holistic representation of group instance information. 
We further visualize the clustering effect of GMP and provide more ablations in supplementary material.

\vspace{-4mm}
\paragraph{Cascade TGA.} \label{exp:cascade}

In Figure \ref{fig:single-vs-cascade}, we evaluate DBNetpp with single TGA and Cascade TGA, respectively, at various training stages with the frozen text detector. 
The results indicate that introducing Cascade TGA not only accelerates convergence, but also produces superior results compared to single TGA.
This structural prior introduced by Cascade TGA reduces the problem of clustering words into paragraphs to a two-stage problem: first clustering words into lines and then clustering lines into paragraphs.

\vspace{-1mm}
\subsection{Qualitative Results}
\vspace{-1mm}

We compare visualizations of generated layouts between Unified Detector, our line-based MaskDINO with TGA and our word-based DeepSolo with TGA in Figure~\ref{fig:visualization}. 
We observe that our line-based model performs better in the details, with more complete text masks and accurate grouping results. 
It is also noteworthy that our DeepSolo with TGA simultaneously produces the result of text detection, recognition, and layout analysis as a unified model.
Facing more challenging in word-based layout analysis, it shows slight defects like losing the capture of small-size texts.

\vspace{-2mm}
\section{Conclusion}
\vspace{-1mm}
We present Text Grouping Adapter (TGA), a versatile module that enhances the capability of various pre-trained text detectors to serve for layout analysis. TGA takes text masks and image features as inputs to predict text group masks from text instance features. It facilitates the full exploitation of well-trained text detectors and easily obtainable text detection data. This work provides insights and a practical solution for aligning layout analysis with text detection and also has the potential to model general object relations.

\vspace{-2mm}
\section*{Acknowledgements}
\vspace{-1mm}
Tianci Bi and Nanning Zheng were supported in part by
NSFC under grant No. 62088102.

{
    \small
    \bibliographystyle{ieeenat_fullname}
    \bibliography{main}
}


\end{document}